\documentclass{article}
\usepackage{ijcai19}
\usepackage{natbib}
\usepackage{times}
\usepackage{soul}
\usepackage{url}
\usepackage[hidelinks]{hyperref}
\usepackage[utf8]{inputenc}
\usepackage[small]{caption}
\usepackage{graphicx}
\usepackage{mathrsfs}
\usepackage{multirow}
\usepackage{amssymb}
\usepackage{amsmath}
\usepackage{booktabs}
\usepackage{algorithm}
\usepackage{algorithmic}
\usepackage{xspace}
\usepackage{wrapfig}
\usepackage{enumitem}

\title{A Variational Dirichlet Framework for Out-of-Distribution Detection}

\author{Wenhu Chen$^{\dagger}$, Yilin Shen$^{\ddag}$, Hongxia Jin$^{\ddag}$, William Wang$^{\dagger}$\\
\affiliations
University of California, Santa Barbara$^{\dagger}$\\
Samsung Research, Moutain View$^{\ddag}$\\
{\tt\small \{wenhuchen,william\}@cs.ucsb.edu \{yilin.shen, hongxia.jin\}@samsung.com}}

\newcommand{\loss}{\mathcal{L}^{VI}}
\newcommand{\dloss}{\mathcal{L}^{D}}
\newcommand{\expect}[1]{\ensuremath{\underset{#1}{\mathbb{E}}\xspace}}
\newcommand{\true}{\ensuremath{p(z|y)}}
\DeclareMathOperator*{\argmax}{arg\,max}

\begin{document}

\maketitle

\begin{abstract}
With the recently rapid development in deep learning, deep neural networks have been widely adopted in many real-life applications. However, deep neural networks are also known to have very little control over its uncertainty for unseen examples, which potentially causes very harmful and annoying consequences in practical scenarios. In this paper, we are particularly interested in designing a higher-order uncertainty metric for deep neural networks and investigate its effectiveness under the out-of-distribution detection task proposed by~\cite{hendrycks2016baseline}. Our method first assumes there exists an underlying higher-order distribution $\mathbb{P}(z)$, which controls label-wise categorical distribution $\mathbb{P}(y)$ over classes on the K-dimension simplex, and then approximate such higher-order distribution via parameterized posterior function $p_{\theta}(z|x)$ under variational inference framework, finally we use the entropy of learned posterior distribution $p_{\theta}(z|x)$ as uncertainty measure to detect out-of-distribution examples. Further, we propose an auxiliary objective function to discriminate against synthesized adversarial examples to further increase the robustness of the proposed uncertainty measure. Through comprehensive experiments on various datasets, our proposed framework is demonstrated to consistently outperform competing algorithms.
\end{abstract}

\begin{figure*}[thb]
    \centering
    \includegraphics[width=1.0\linewidth]{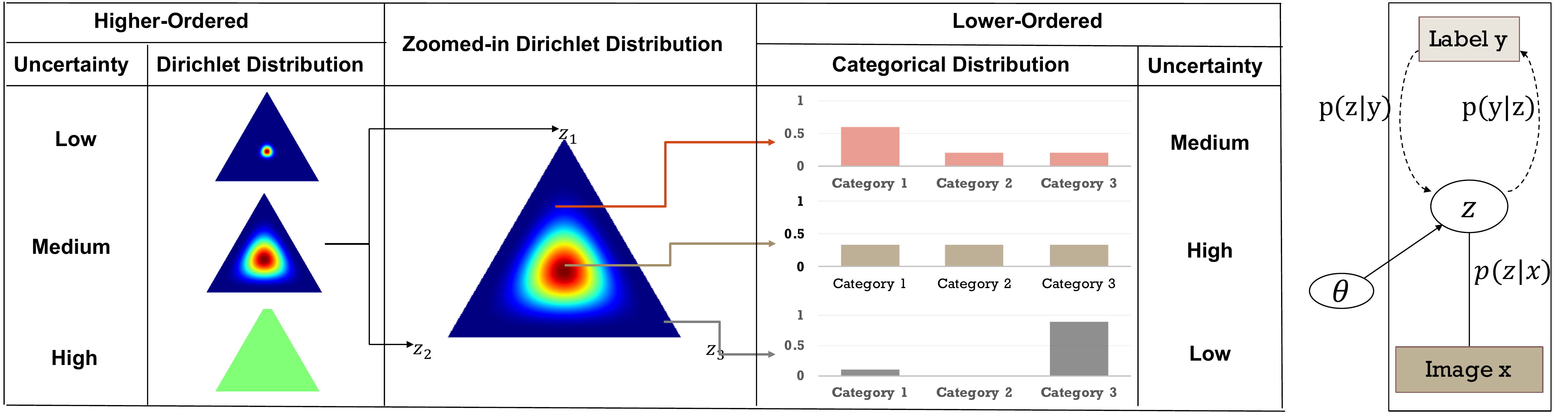}
    \caption{An intuitive explanation of higher-order distribution and lower-order distribution and their uncertainty measures.}
    \label{fig:higher-lower}
\end{figure*}
\section{Introduction}
Recently, deep neural networks~\citep{lecun2015deep} have surged and replaced the traditional machine learning algorithms to demonstrate its potentials in many real-life applications like image classification~\citep{deng2009imagenet,he2016deep}, and machine translation~\citep{wu2016google,vaswani2017attention}, etc. However, unlike the traditional machine learning algorithms like Gaussian Process, Logistic Regression, etc, deep neural networks are very limited in their capability to measure the uncertainty over the unseen cases and tend to produce over-confident predictions. Such an over-confidence issue is known to be harmful or offensive in real-life applications. Even worse, such models are prone to adversarial attacks and raise concerns in AI safety~\citep{DBLP:journals/corr/GoodfellowSS14}. Therefore, it is very essential to design a robust and accurate uncertainty metric in deep neural networks in order to better deploy them into real-world applications to benefit human beings. Recently, An out-of-distribution detection task was proposed in~\cite{hendrycks2016baseline} as a benchmark to promote the uncertainty research in the deep learning community. In the baseline approach, the highest softmax probability is adopted directly as the indicator for the model's confidence to distinguish in- from out-of-distribution data. Later on, many follow-up algorithms~\citep{liang2017enhancing,lee2017training,shalev2018out,devries2018learning} have been proposed to achieve better performance on this benchmark. In ODIN~\citep{liang2017enhancing}, the authors follow the temperature scaling and input perturbation algorithms~\citep{pereyra2017regularizing,balan2015bayesian} to widen the distance between in- and out-of-distribution examples. Later on, adversarial training~\citep{lee2017training} is introduced to explicitly introduce boundary examples as negative training data to help increase the model's robustness. In ~\cite{devries2018learning}, the authors proposed to directly output a real value between [0, 1] as the confidence measure. A recent paper~\citep{shalev2018out} leverages the semantic dense representation into the target labels to better separate the label space and uses the cosine similarity score as the confidence measure for out-of-distribution detection. The state-of-the-art results are achieved by Mahalanobis algorithm~\citep{lee2018simple}, which incorporates the low-level features of deep neural networks as ensemble to discriminate between in- and out-of-distribution data. 

These methods though achieve significant results on out-of-distribution detection tasks, conflate different levels of uncertainty as pointed in~\cite{malinin2018predictive}. For example, when presented with two pictures, one is a forged by mixing dog, cat and horse pictures, the other is a real but unseen dog, the model might predict same belief for these two inputs as \{cat:33.3\%, dog:33.3\%, horse:33.3\%\}. Given such faked image, the existing measures like~\cite{liang2017enhancing,shalev2018out,hendrycks2016baseline} will misclassify both images as from out-of-distribution because they are unable to separate the two uncertainty sources: whether the uncertainty is due to the data noise (class overlap) or whether the data is far from the manifold of training data. More specifically, they fail to distinguish between the lower-order (aleatoric) uncertainty~\citep{gal2016uncertainty}, and higher-order (episdemic) uncertainty~\cite{gal2016uncertainty}, which leads to their inferior performances in detecting out-domain examples. 

In order to resolve the issues presented by lower-order uncertainty measures, we are motivated to design an effective higher-order uncertainty measure for out-of-distribution detection. Inspired by Subjective Logic~\citep{DBLP:books/sp/Josang16,yager2008classic,sensoy2018evidential}, we first interpret the label-wise categorical distribution $\mathbb{P}(y)$ as a K-dimensional variable $z$ generated from a higher-order distribution $\mathbb{P}(z)$ over the simplex $\mathbb{S}_k$, and then study the higher-order uncertainty by investigating the statistical properties of such underlying distribution. Under a Bayesian framework with data pair $(x, y) \in D$, we propose to use variational inference to approximate such ``true" latent distribution $\mathbb{P}(z)=p(z|y)$ by a neural-network parameterized Dirichlet posterior $p_{\theta}(z|x)$. Under the given Bayesian framework, we propose to use its entropy $\mathbb{H}(p_{\theta}(z|x))$ for out-of-distribution detection. To further enhance the proposed uncertainty metric, we design an auxiliary discriminative objective to help the Dirichlet model discriminate in-domain data from synthesized adversarial data generated using fast-sign gradient method (FGSM)~\citep{kurakin2018adversarial}, which is combined with the variational evidence lower bound to build a unified objective to train the parameterized posterior function. Finally, we compute the entropy of the approximated Dirichlet distribution as the uncertainty measure to decide which source distribution a given image is from. To sum up, the contributions of this paper are described as follows:
\begin{itemize}
    \item We propose a variational Dirichlet algorithm for deep neural network classification problem and define a higher-order uncertainty measure.
    \item We further design the discriminative function to enhance the proposed uncertainty measure and achieve significant performance under different datasets and deep neural architectures. 
\end{itemize}

%\begin{figure}[thb]
%\centering
%\includegraphics[width=1.0\linewidth]{latent}
%\caption{Illustration of the Plate Notation of our Variation Model}\label{fig:latent}
%\end{figure}
\section{Model}
In this paper, we particularly consider the image classification problem with image input as $x$ and label as $y$. By interpreting the label-level categorical distribution $\mathbb{P}(y)=[p(y=\omega_1), \cdots, p(y=\omega_k)]$ as a random variable $z=\{z \in \mathbb{R}^k:\sum_{i=1}^k z_i=1\}$ lying on a K-dimensional simplex $\mathbb{S}_k$, we assume there exists an underlying higher-order distribution $\mathcal{P}(z)$ controlling the generation of $z$. As depicted in the LHS of~\autoref{fig:higher-lower}, each point on the simplex $\mathbb{S}_k$ is itself a unique categorical distribution $\mathbb{P}(y)$ over different categories. The high-order distribution $\mathbb{P}(z)$ represents the underlying generation function of the simplex $\mathbb{S}_k$. By studying the statistical properties (entropy, mutual information, etc) of $\mathbb{P}(z)$, we can quantitatively study its higher-order uncertainty. 
\paragraph{Variational Bayesian}
Here we consider a Bayesian inference framework with a given dataset $(x, y) \in D$ and show its plate notation in the RHS of~\autoref{fig:higher-lower}, where $x$ denotes the observed variable (images), $y$ is the groundtruth label (known at training but unknown as testing), and $z$ is latent variable higher-order variable. We assume that the ``true" posterior distribution is encapsulated in the partially observable groundtruth label $y$, thus it can be written as $\mathbb{P}(z)=p(z|y)$. During test time, due to the inaccessibility of $y$, we need to approximate such ``true" underlying distribution with a given input image $x$. Therefore, we propose to parameterize a posterior model $p_{\theta}(z|x)$ and optimize its parameters $\theta$ to approach such ``true" posterior $p(z|y)$ by minimizing their KL-divergence $D_{KL}(p_{\theta}(z|x)||p(z|y))$. With the parameterized posterior $p_{\theta}(z|x)$, we are able to infer the higher-order distribution over $z$ given an unseen image $x^*$ and quantitatively to estimate its higher-order uncertainty.

\begin{figure}[thb]
    \centering
    \includegraphics[width=1.0\linewidth]{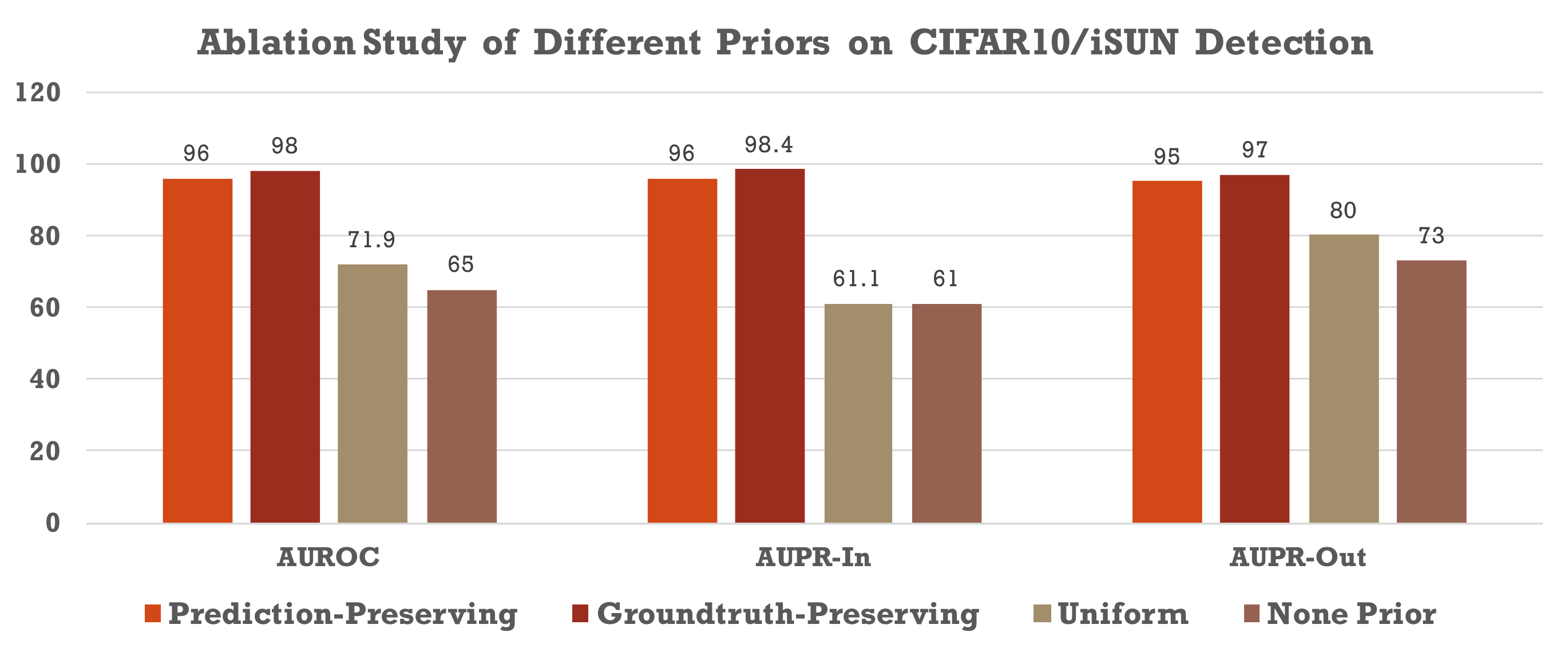}
    \caption{An intuitive explanation of different prior functions.}
    \label{fig:priors_funcs}
\end{figure}

In order to minimize the KL-divergence, we leverage the variational inference framework to decompose it into two components as follows:
\begin{align}
\begin{split}
    &D_{KL}(p_{\theta}(z|x)||\mathbb{P}(z))\\
    =& \int_z p_{\theta}(z|x)\log \frac{p_{\theta}(z|x)}{p(z|y)} dz\\
    =& \int_z p_{\theta}(z|x)\log \frac{p_{\theta}(z|x)p(y)}{p(z, y)} dz\\
    =& \int_z p_{\theta}(z|x)\log \frac{p_{\theta}(z|x)}{p(z, y)} dz + \log p(y)\\
    =& \int_z p_{\theta}(z|x)\log \frac{p_{\theta}(z|x)}{p(z)p(y|z)} dz + \log p(y)\\
\end{split}
\end{align}
We further divide $\frac{p_{\theta}(z|x)}{p(z)p(y|z)}$ into two components as follows:
\begin{align*}
    &D_{KL}(p_{\theta}(z|x)||\mathbb{P}(z))\\
    =& \int_z p_{\theta}(z|x)\log \frac{p_{\theta}(z|x)}{p(z)}dz \\
     &- \int_z p_{\theta}(z|x) \log p(y|z)dz + \log p(y)\\
    =& KL(p_{\theta}(z|x)||p(z)) -\expect{z \sim p_{\theta}(z|x)} [\log p(y|z)] + \log p(y)\\
    =& -[\expect{z \sim p_{\theta}(z|x)} [\log p(y|z)] - KL(p_{\theta}(z|x)||p(z))] + \log p(y)\\
    =& -\loss(\theta) + \log p(y)    
\end{align*}
%\begin{align}
%    D_{KL}(p_{\theta}(z|x)||p(z|y)) =  -\loss(\theta) + \log p(y)
%\end{align}
where $\loss(\theta)$ is better known as the variational evidence lower bound, and $\log p(y)$ is the marginal log likelihood over the ground truth label $y$. 
\begin{align}
    \loss(\theta) = \expect{z \sim p_{\theta}(z|x)}[\log p(y|z)] - D_{KL}(p_{\theta}(z|x)||p(z))
\end{align}
Since the marginal distribution $p(y)$ is constant w.r.t $\theta$, minimizing the KL-divergence $D_{KL}(p_{\theta}(z|x)||\true)$ is equivalent to maximizing the evidence lower bound $\loss(\theta)$. 
\paragraph{Dirichlet Parameterization}
Here we propose to use Dirichlet family to realize the higher-order distribution $p_{\theta}(z|x) = Dir(z|\alpha)$ due to its tractable analytical properties. The probability density function of Dirichlet distribution over all possible values of the K-dimensional stochastic variable $z$ can be written as:
\begin{align}
    Dir(z|\alpha) = \begin{cases} 
    \frac{1}{B(\alpha)}\prod_{i=1}^K z_i^{\alpha_i - 1} \quad &for \quad z \in \mathbb{S}_k\\
    0 \qquad & otherwise,
    \end{cases}
\end{align}
where $\alpha$ is the concentration parameter of the Dirichlet distribution and $B(\alpha)=\frac{\prod_i^K \Gamma(\alpha_i)}{\Gamma(\sum_i^k \alpha_i)}$ is the normalization factor. Since the LHS (expectation of log probability) has a closed-formed solution, we can rewrite the empirical lower bound on given dataset $D$ as follows:
\begin{align}
\small
\begin{split}
    \loss(\theta) = \sum_{(x, y) \in D}  [\psi(\alpha_y) - \psi(\alpha_0) - D_{KL}(Dir(z|\alpha)||p(z))]
\end{split}
\end{align}
where $\alpha_0$ is the sum of concentration parameter $\alpha$ over K dimensions. However, it is in general difficult to select a perfect model prior to craft a model posterior which induces an the distribution with the desired properties. Here, we assume the prior distribution is as Dirichlet distribution $Dir(\hat{\alpha})$ with concentration parameters $\hat{\alpha}$ and specifically talk about three intuitive prior functions in~\autoref{fig:priors_funcs}.
The first uniform prior aggressively pushes all dimensions towards 1, while the *-preserving priors are less strict by allowing one dimension of freedom in the posterior concentration parameter $\alpha$. This is realized by copying the value from a certain dimension of posterior concentration parameter $\alpha$ to the uniform concentration to unbind $\alpha_k$ from KL-divergence computation. Given the prior concentration parameter $\hat{\alpha}$, we can obtain a closed-form solution for the evidence lower bound as follows:
\begin{align}
\small
\begin{split}
     \loss(\theta) = &\sum_{(x, y) \in D}  [\psi(\alpha_y) - \psi(\alpha_0) - \log \frac{B(\hat{\alpha})}{B(\alpha)} - \\
                    &\sum_{i=1}^k ({\alpha}_i - \hat{\alpha}_i)(\psi(\alpha_i) - \psi(\alpha_0))   
\end{split}
\end{align}
$\Gamma$ denotes the gamma function, $\psi$ denotes the digamma function. In practice, we parameterize $Dir(z|\alpha)$ via a neural network $\alpha = f_{\theta}(x)$ with parameters $\theta$. 
\paragraph{Uncertainty Measure}
Here we propose a to use entropy $\mathbb{H}(p_{\theta}(z|\alpha))$ as the higher-order uncertainty measure. Formally, we write the confidence metric $C(\alpha)$ as follows:
\begin{align}
\small
\begin{split}
    C(\alpha) &= -\mathbb{H}(p_{\theta}(z|\alpha)) = -\int_{z} z Dir(z|\alpha)dz \\
              &= \log B(\alpha) + (\alpha_0 - K) \psi(\alpha_0) - \sum_i^k (\alpha_i - 1)\psi(\alpha_i)
\end{split}
\end{align}
where $\alpha$ is estimated via the deep neural network $f_{\theta}(x)$. Here we use negative entropy as the confidence measure $C(\alpha)$.
\begin{figure}[thb]
\centering
\includegraphics[width=0.9\linewidth]{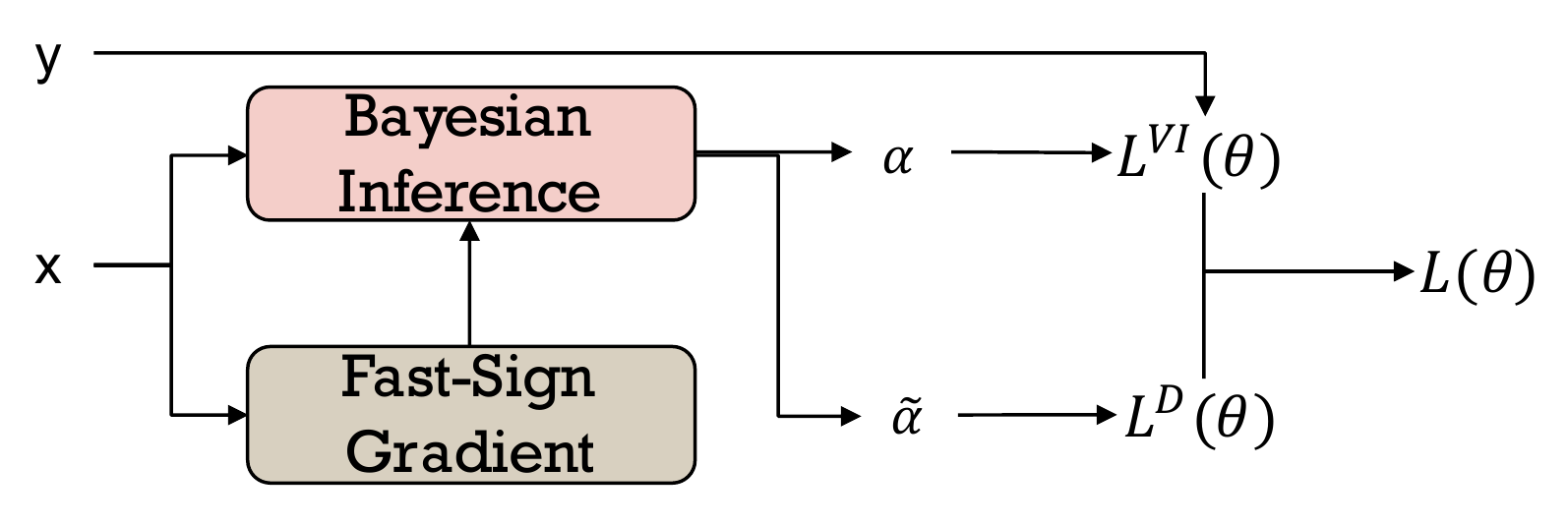}
\caption{Discriminative Training Procedure}
\label{fig:discriminative}
\end{figure}
\paragraph{Discriminative objective}
In order to further enhance the model's robustness against the out-domain examples, we propose a auxiliary discriminative objective $\dloss(\theta)$ to defend against the adversarial examples $\tilde{x}$ generated by fast gradient sign method (FGSM)~\citep{kurakin2018adversarial} as follows:
\begin{align}
\small
\begin{split}
    \dloss(\theta) =& \expect{(x, *) \sim D} [C(f_{\theta}(x))] - \expect{\tilde{x} \sim FGSM(x)} [C(f_{\theta}(\tilde{x}))] \\
    &FGSM: \hat{x} = x - \epsilon sign(\nabla_{x}J(x, y))
\end{split}
\end{align}
Our hope is to that fast-sign adversarial examples can generalize to broader out-of-distribution data sources, so that the model can learn to widen the distance between the in- and out-of-distribution examples.
\paragraph{Optimization}
The auxiliary discriminative function is combined with the Bayesian inference objective to arrive in an interpolation using a balancing factor $\lambda$ as our ultimate objective function $\mathcal{L}(\theta)$:
\begin{align}
    \mathcal{L}(\theta) = \loss(\theta) + \lambda \dloss(\theta)
\end{align}
\begin{table*}[thb]
\small
\centering
\begin{tabular}{lllll} 
\toprule
Data Source                          & Dataset       & Content (classes)                                & \#Train & \#Test  \\ 
\midrule
\multirow{2}{*}{In-Distribution}     & CIFAR-10~\cite{krizhevsky2009learning}      & 10 classes: Airplane, Truck, Bird, etc.          & 60,000  & 10,000  \\
                                     & CIFAR-100~\cite{krizhevsky2009learning}     & 100 classes: Mammals,~ Fish, Flower, etc         & 60,000  & 10,000  \\ 
\midrule
\multirow{4}{*}{Out-of-Distribution} & iSUN~\cite{xiao2010sun}          & 908 clases: Airport, Abby, etc                   & -       & 8,925   \\
                                     & LSUN~\cite{yu2015lsun}          & 10 classes: Bedrooms, Churches, etc              & -       & 10,000  \\
                                     & Tiny-ImageNet~\cite{deng2009imagenet} & 1000 classes: Plant, Natural object, Sports, etc & -       & 10,000  \\
                                     & SVHN~\cite{netzer2011reading}          & 10 classes: The Street View House Numbers        & -       & 26,032  \\
\bottomrule
\end{tabular}
\caption{Overview of in- and out-of-distribution datasets}
\label{tab:dataset}
\end{table*}
We adopt gradient ascent to optimize $\mathcal{L}(\theta)$ against the parameters $\theta$. Specifically, we write the derivative of $\loss(\theta)$ and $\dloss(\theta)$ w.r.t to parameters $\theta$ based on the chain-rule: $\frac{\partial \loss}{\partial \theta} = \frac{\partial \loss}{\partial \alpha} \odot \alpha \boldsymbol{\cdot} \frac{\partial f_{\theta}(x)}{\partial \theta}$, where $\odot$ is the Hardamard product and $\frac{\partial f_{\theta}(x)}{\partial \theta}$ is the Jacobian matrix.
\begin{align}
\small
\begin{split}
    \frac{\partial \loss}{\partial \alpha} = \sum_{(x, y) \in B(x, y)} [&\frac{\partial [\psi(\alpha_y) - \psi(\alpha_0)]}{\partial \alpha}  + \\
                            &\frac{\partial D_{KL}(Dir(\alpha)||Dir(\hat{\alpha}))}{\partial \alpha}]\\
    \frac{\partial \dloss}{\partial \alpha} = \expect{(x, *) \sim B(x, y)} &[\frac{\partial C(\alpha)}{\partial \alpha}] - \expect{\tilde{x} \sim FGSM(x)} [\frac{\partial C(\tilde{\alpha})}{\partial \tilde{\alpha}}]
\end{split}
\end{align}
where $B(x,y)$ denotes a mini-batch sampled from dataset $D$. During inference time, we use the marginal probability of assigning input $x$ to label $i$ as the classification evidence:
\begin{equation}
    p(y=i|x)=\int_{z}p(y=i|z)p_{\theta}(z|x)  dz= \frac{\alpha_i}{\sum_{j=1}^k \alpha_j}
\end{equation}
Therefore, we can use the maximum $\alpha$'s index as the model prediction class during inference $\hat{y} = \argmax_i p(y=i|x) = \argmax_i \alpha_i$.
\paragraph{OOD Detection} For each input $x$, we first feed it into neural network $f_{\theta}(x)$ to compute the concentration parameters $\alpha$. Specifically, we compare the confidence $C(\alpha)$ to the threshold $T$ and say that the data $x$ follows in-distribution if the confidence score $C(\alpha)$ is above the threshold and that the data $x$ follows out-of-distribution, otherwise. 

\section{Experiments}
In order to evaluate our variational Dirichlet method on out-of-distribution detection, we follow the previous paper~\citep{hendrycks2016baseline,liang2017enhancing} to replicate their experimental setup. Throughout our experiments, a neural network is trained on some in-distribution datasets to distinguish against the out-of-distribution examples represented by images from a variety of unrelated datasets. For each sample fed into the neural network, we will calculate the confidence metric to determine which distribution the sample comes from. Finally, several different evaluation metrics are used to measure and compare how well different detection methods can separate the two distributions.

\subsection{Training Details}
Here we list all the datasets used in~\autoref{tab:dataset}, which are available in github\footnote{\url{https://github.com/ShiyuLiang/odin-pytorch}}. For CIFAR10 and CIFAR100 dataset, we separately use the widely adopted network architectures VGG13~\cite{simonyan2014very}, ResNet18~\citep{he2016deep}, ResNet34~\citep{he2016deep}, Wide-ResNet~\citep{DBLP:conf/bmvc/ZagoruykoK16} and ResNeXt~\citep{xie2017aggregated} with publicly available code\footnote{\url{https://github.com/bearpaw/pytorch-classification}}. We use the publicly available implementation in github\footnote{\url{https://github.com/1Konny/FGSM}} to implement FGSM~\citep{kurakin2018adversarial} for generating adversarial examples (perturbation magnitude $\epsilon$ is set to  0.2). Our method is implemented with Pytorch library. All models are trained using stochastic gradient descent with Nesterov momentum of 0.9, and weight decay with 5e-4. We train all models for 200 epochs with a 128 batch size. We initialize the learning with 0.1 and reduced by a factor of 5 at 60th, 120th and 180th epochs. We cut off the gradient norm by 1 to prevent from potential gradient exploding error, after training, after the classification accuracy on the validation set converges and we use the saved model for out-of-distribution detection. The best reported results are all using groundtruth-preserving prior to compute the evidence lower bound, the balancing factor of $\lambda=0.1$ is adopted for our experimental settings.

\begin{table}
\small
\centering
\begin{tabular}{c|cccc} 
\hline
\multirow{2}{*}{Dataset} & \multicolumn{4}{c}{Cross-Entropy/Ours} \\
\cline{2-5}
                       & VGG13 & ResNet-18 & ResNet-34 & WideResNet\\
\hline
C-10                & 93.3/93.1  & 94.4/94.0       & -  & - \\
C-100               & -  &  -              & 79.4/79.1  & 80.5/80.7 \\ 
\hline
\end{tabular}
\caption{Classification accuracy of Dirichlet framework on CIFAR10/100 under different architectures.}
\label{tab:accuracy}
\end{table}

\begin{table}[thb]
\small
\centering
\begin{tabular}{lllll} 
\hline
\multirow{2}{*}{\begin{tabular}[c]{@{}l@{}}IND/OOD\\Model\end{tabular}}                     & \multirow{2}{*}{Method} & FPR@           & \multirow{2}{*}{\begin{tabular}[c]{@{}l@{}}Detection \\Error \end{tabular}} & \multirow{2}{*}{AUROC}  \\
                                                                                            &                         & TPR95          &                                                                             &                         \\ 
\hline
\multirow{5}{*}{\begin{tabular}[c]{@{}l@{}}CIFAR10/\\iSUN\\~\\VGG13\end{tabular}}           & Baseline                & 43.8           & 11.4                                                                        & 94                      \\
                                                                                            & ODIN                    & 22.4           & 10.2                                                                        & 95.8                    \\
                                                                                            & Confidence              & 16.3           & 8.5                                                                         & 97.5                    \\
                                                                                            & Semantic                & 23.2           & 10.2                                                                        & 96.4                    \\
                                                                                            & Ours                    & \textbf{10.7 } & \textbf{7.4 }                                                               & \textbf{97.7 }          \\ 
\hline
\multirow{5}{*}{\begin{tabular}[c]{@{}l@{}}CIFAR10/\\LSUN\\~\\VGG13\end{tabular}}           & Baseline                & 41.9           & 11.5                                                                        & 94                      \\
                                                                                            & ODIN                    & 20.2           & 9.8                                                                         & 95.9                    \\
                                                                                            & Confidence              & 16.4           & 8.3                                                                         & 97.5                    \\
                                                                                            & Semantic                & 22.9           & 13.9                                                                        & 96.0                    \\
                                                                                            & Ours                    & \textbf{10.3 } & \textbf{7.4 }                                                               & \textbf{97.8 }          \\ 
\hline
\multirow{5}{*}{\begin{tabular}[c]{@{}l@{}}CIFAR10/\\Tiny-ImgNet\\~\\VGG13\end{tabular}} & Baseline                & 43.8           & 12                                                                          & 93.5                    \\
                                                                                            & ODIN                    & 24.3           & 11.3                                                                        & 95.7                    \\
                                                                                            & Confidence              & 18.4           & 9.4                                                                         & 97                      \\
                                                                                            & Semantic                & 19.8           & 10.1                                                                        & 96.5                    \\
                                                                                            & Ours                    & \textbf{13.8 } & \textbf{7.9 }                                                               & \textbf{97.5 }          \\
\hline
\end{tabular}
\caption{Experimental Results on VGG13~\citep{simonyan2014very} architecture, where Confidence refers to~\cite{devries2018learning} and Semantic refers to~\cite{shalev2018out}, most results are copied from original paper.}
\label{tab:vgg}
\end{table}

\begin{table*}[thb]
\small
\centering
\begin{tabular}{lllccccc} 
\toprule
Model                                                                           & OOD & Method         & FPR@TPR95                 & Detection Error           & AUROC                     & AUPR~ In                  & AUPR Out                   \\ 
\midrule
\multirow{3}{*}{\begin{tabular}[c]{@{}l@{}}WideResNet\\CIFAR-100 \end{tabular}} & iSUN                                                   & Base/ODIN/Ours & 82.7/57.3/\textbf{18.0}  & 43.9/31.1/\textbf{11.1} & 72.8/86.6/\textbf{95.5} & 74.2/85.9/\textbf{95.5} & 69.2/84.9/\textbf{95.6}  \\
                                                                                & LSUN                                                   & Base/ODIN/Ours & 82.2/56.5/\textbf{13.3} & 43.6/30.8/\textbf{8.8}  & 73.9/86.0/\textbf{97.1} & 75.7/86.2/\textbf{97.1}  & 69.2/84.9/\textbf{97.2}  \\
                                                                                & TinyImgNet                                               & Base/ODIN/Ours & 79.2/55.9/\textbf{16.8} & 42.1/30.4/\textbf{10.2} & 72.2/84.0/\textbf{96.2}  & 70.4/82.8/\textbf{95.7} & 70.8/84.4/\textbf{96.4}   \\ 
\midrule
\multirow{3}{*}{\begin{tabular}[c]{@{}l@{}}ResNeXt-29\\CIFAR-100 \end{tabular}} & iSUN                                                   & Base/ODIN/Ours & 82.2/61.6/\textbf{18.4}  & 31.0/21.4/\textbf{11.2} & 74.5/86.4/\textbf{94.9} & 79.8/89.1/\textbf{95.3} & 67.7/82.7/\textbf{94.0}   \\
                                                                                & LSUN                                                   & Base/ODIN/Ours & 82.2/62.4/\textbf{13.6}  & 31.8/22.1/\textbf{8.7}  & 73.6/85.9/\textbf{96.5} & 77.4/87.8/\textbf{96.8} & 69.5/83.9/\textbf{95.8}  \\
                                                                                & TinyImgNet                                               & Base/ODIN/Ours & 79.6/60.2/\textbf{16.9} & 31.0/21.5/\textbf{9.9}   & 75.1/86.5/\textbf{96.5} & 78.4/88.2/\textbf{96.8} & 71.6/84.8/\textbf{95.8}  \\
\bottomrule
\end{tabular}
\caption{More experimental results on CIFAR100 dataset on WideResNet and ResNeXt architeture.}
\label{tab:more_100}
\end{table*}

\begin{table}[thb]
\small
\centering
\begin{tabular}{lllll} 
\midrule
\begin{tabular}[c]{@{}l@{}} IND/OOD\\Model \end{tabular}                                      & Method      & \begin{tabular}[c]{@{}l@{}} FPR@\\TPR95 \end{tabular} & \begin{tabular}[c]{@{}l@{}} Detection\\Error \end{tabular} & AUROC           \\ 
\midrule
\multirow{5}{*}{\begin{tabular}[c]{@{}l@{}} CIFAR10/\\Tiny-ImgNet\\~\\ResNet18 \end{tabular}} & Baseline    & 59.0                                                  & 15.1                                                       & 91.1            \\
                                                                                              & ODIN        & 32.1                                                  & 11.2                                                       & 94.9            \\
                                                                                              & DPN         & 28.4                                                     & 13.6                                                          & 93.0            \\
                                                                                              & Mahalanobis & \textbf{2.9}                                          & \textbf{0.6}                                               & 96.3            \\
                                                                                              & Ours        & 17.1                                                  & 8.7                                                        & \textbf{96.8}   \\ 
\midrule
\multirow{5}{*}{\begin{tabular}[c]{@{}l@{}} CIFAR10/\\LSUN\\~\\ResNet18 \end{tabular}}        & Baseline    & 50.2                                                  & 12.3                                                       & 93.1            \\
                                                                                              & ODIN        & 17.9                                                  & 8.4                                                        & 96.9            \\
                                                                                              & DPN         & 57.4                                                     & 20.5                                                          & 90.2            \\
                                                                                              & Mahalanobis & \textbf{1.2}                                          & \textbf{0.3}                                               & 97.5            \\
                                                                                              & Ours        & 7.7                                                   & 5.9                                                        & \textbf{98.3}   \\ 
\midrule
\multirow{5}{*}{\begin{tabular}[c]{@{}l@{}} CIFAR10/\\SVHN\\~\\ResNet18 \end{tabular}}        & Baseline    & 49.5                                                  & 13.3                                                       & 92.0            \\
                                                                                              & ODIN        & 29.7                                                  & 15.1                                                       & 91.7            \\
                                                                                              & DPN         & 20.1                                                     & 12.7                                                          & \textbf{95.9}   \\
                                                                                              & Mahalanobis & \textbf{12.2}                                         & \textbf{2.3}                                               & 92.6            \\
                                                                                              & Ours        & 28.7                                                  & 13.6                                                       & 93.2            \\
\midrule
\end{tabular}
\caption{Experimental Results on ResNet18~\citep{he2016deep} architecture, where Mahalanobis refers to~\cite{lee2018simple} and DPN refers to~\cite{malinin2018predictive}, most results are copied from the original paper.}
\label{tab:resnet_10}
\end{table}

\begin{table}[thb]
\small
\centering
\begin{tabular}{lllll} 
\toprule
\begin{tabular}[c]{@{}l@{}} IND/OOD\\Model \end{tabular}                                       & Method      & \begin{tabular}[c]{@{}l@{}} FPR@\\TPR95 \end{tabular} & \begin{tabular}[c]{@{}l@{}} Detection\\Error \end{tabular} & AUROC             \\ 
\midrule
\multirow{4}{*}{\begin{tabular}[c]{@{}l@{}} CIFAR-100/\\iSUN\\~\\ResNet34 \end{tabular}}       & ODIN        & 61.3                                                  & 23.7                                                       & 83.6              \\
                                                                                               & Semantic    & 58.4                                                  & 21.4                                                       & 85.2              \\
                                                                                               & Mahalanobis & \textbf{18.7}                                       & \textbf{11.6}                                            & 94.1              \\
                                                                                               & Ours        & 19.8                                                  & 12.2                                                       & \textbf{94.2}   \\ 
\midrule
\multirow{4}{*}{\begin{tabular}[c]{@{}l@{}} CIFAR-100/\\LSUN\\~\\ResNet34 \end{tabular}}       & ODIN        & 76.8                                                  & 42.4                                                       & 78.9              \\
                                                                                               & Semantic    & 79.5                                                  & 42.2                                                       & 79.0              \\
                                                                                               & Mahalanobis & 14.9                                                  & \textbf{9.0}                                             & 95.4              \\
                                                                                               & Ours        & \textbf{14.5}                                       & 9.6                                                        & \textbf{95.9}   \\ 
\midrule
\multirow{4}{*}{\begin{tabular}[c]{@{}l@{}}CIFAR-100/\\Tiny-ImgNet\\~\\ResNet34 \end{tabular}} & ODIN        & 63.9                                                  & 25.2                                                       & 82.3              \\
                                                                                               & Semantic    & 62.4                                                  & 24.4                                                       & 83.1              \\
                                                                                               & Mahalanobis & \textbf{12.0}                                       & \textbf{9.1}                                             & \textbf{96.5}   \\
                                                                                               & Ours        & 16.3                                                  & 10.3                                                       & 95.3              \\
\bottomrule
\end{tabular}
\caption{Experimental results on ResNet34 architecture on CIFAR100 dataset, where Semantic refers to~\cite{shalev2018out} and Mahalanobis refers to~\cite{lee2018simple}.}
\label{tab:resnet_100}
\end{table}
\subsection{Experimental Results}
We measure the quality of out-of-distribution detection using the established metrics for this task~\citep{hendrycks2016baseline,liang2017enhancing,shalev2018out}.
\begin{enumerate}[leftmargin=0.5cm]
\setlength{\itemsep}{0pt}
\item FPR at 95\% TPR (lower is better): Measures the false positive rate (FPR) when the true positive rate (TPR) is equal to 95\%.
\item Detection Error (lower is better): Measures the minimum possible misclassification probability defined by $\min_{\delta} \{0.5P_{in}(f(x) \leq \delta) + 0.5P_{out}(f(x) > \delta)\}$.
\item AUROC (larger is better): Measures the Area Under the Receiver Operating Characteristic curve. The Receiver Operating Characteristic (ROC) curve plots the relationship between TPR and FPR.
\item AUPR (larger is better): Measures the Area Under the Precision-Recall (PR) curve, where AUPR-In refers to using in-distribution as positive class and AUPR-Out refers to using out-of-distribution as positive class.
\end{enumerate}
Before reporting the out-of-distribution detection results, we first measure the classification accuracy of our proposed method on the two in-distribution datasets in ~\autoref{tab:accuracy}, from which we can observe that our proposed algorithm has minimum impact on the classification accuracy. 

\paragraph{CIFAR10 experiments}
Here we first demonstrate our experimental results on CIFAR10 datasets with VGG13~\citep{simonyan2014very}  (see~\autoref{tab:vgg}) and ResNet18~\citep{he2016deep} (see~\autoref{tab:resnet_10}). In~\autoref{tab:vgg}, we mainly compare against Baseline~\citep{hendrycks2016baseline}, ODIN~\citep{liang2017enhancing}, Confidence~\citep{devries2018learning} and Semantic~\citep{shalev2018out} under the VGG13~\citep{simonyan2014very} architecture. We can easily observe that our proposed method can significantly outperform competing algorithms across all metrics. In~\autoref{tab:resnet_10}, we mainly compare against ODIN~\citep{liang2017enhancing}, DPN~\citep{malinin2018predictive} and Mahalanobis~\citep{lee2018simple} under ResNet18~\citep{he2016deep} architecture. We can observe that the Mahalanobis algorithm performs extremely well on FPR(TRP=95\%) and detection error metrics, but our method is superior in terms of the AUROC metric. 
\paragraph{CIFAR100 experiments}
Here we experiment with the large-scaled CIFAR100 dataset to further investigate the effectiveness of our proposed algorithm. In~\autoref{tab:resnet_100}, we mainly compare against ODIN~\citep{liang2017enhancing}, Semantic~\citep{shalev2018out}, Mahalanobis~\citep{lee2018simple}. under the ResNet34 architecture. We can observe very similar trends as~\autoref{tab:resnet_10} where both our method and Mahalanobis are significantly outperforming the competing algorithms, though the Mahalanobis method achieves very surprising FPR(TPR=95\%) and detection error scores, it lags behind us in terms of AUROC measure. We also report more results in~\autoref{tab:more_100} using other networks like WideResNet~\citep{DBLP:conf/bmvc/ZagoruykoK16} (depth=28, widening factor=10) and ResNext~\citep{xie2017aggregated} (depth=29, widening factor=8).
\begin{figure}[thb]
    \centering
    \includegraphics[width=1.0\linewidth]{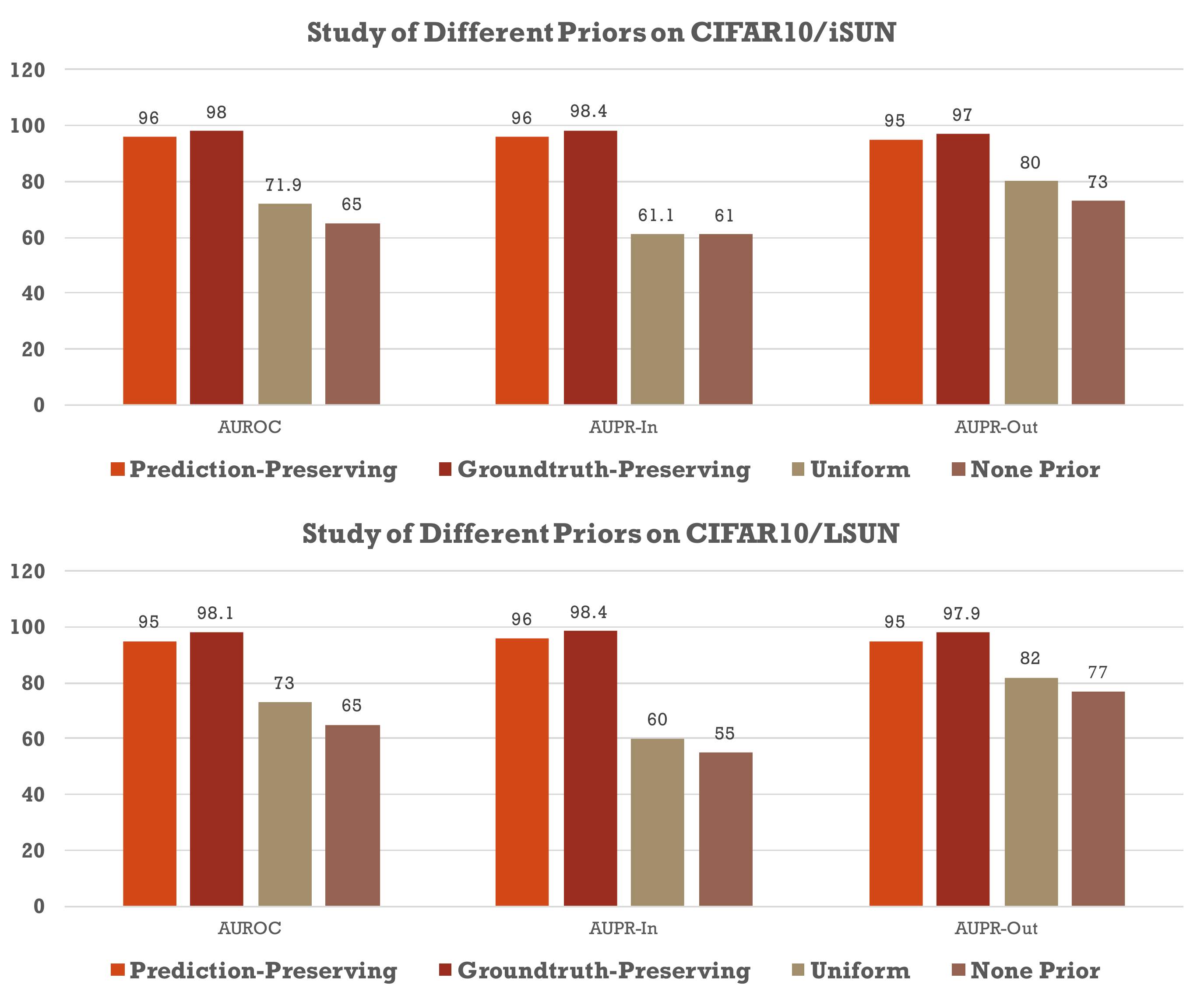}
    \caption{Impact of different prior distributions. The network architecture is VGG13 with CIFAR10 as in-distribution dataset and iSUN/LSUN as out-of-distribution dataset. }
    \label{fig:priors}
\end{figure}
\begin{figure}[thb]
    \centering
    \includegraphics[width=1.0\linewidth]{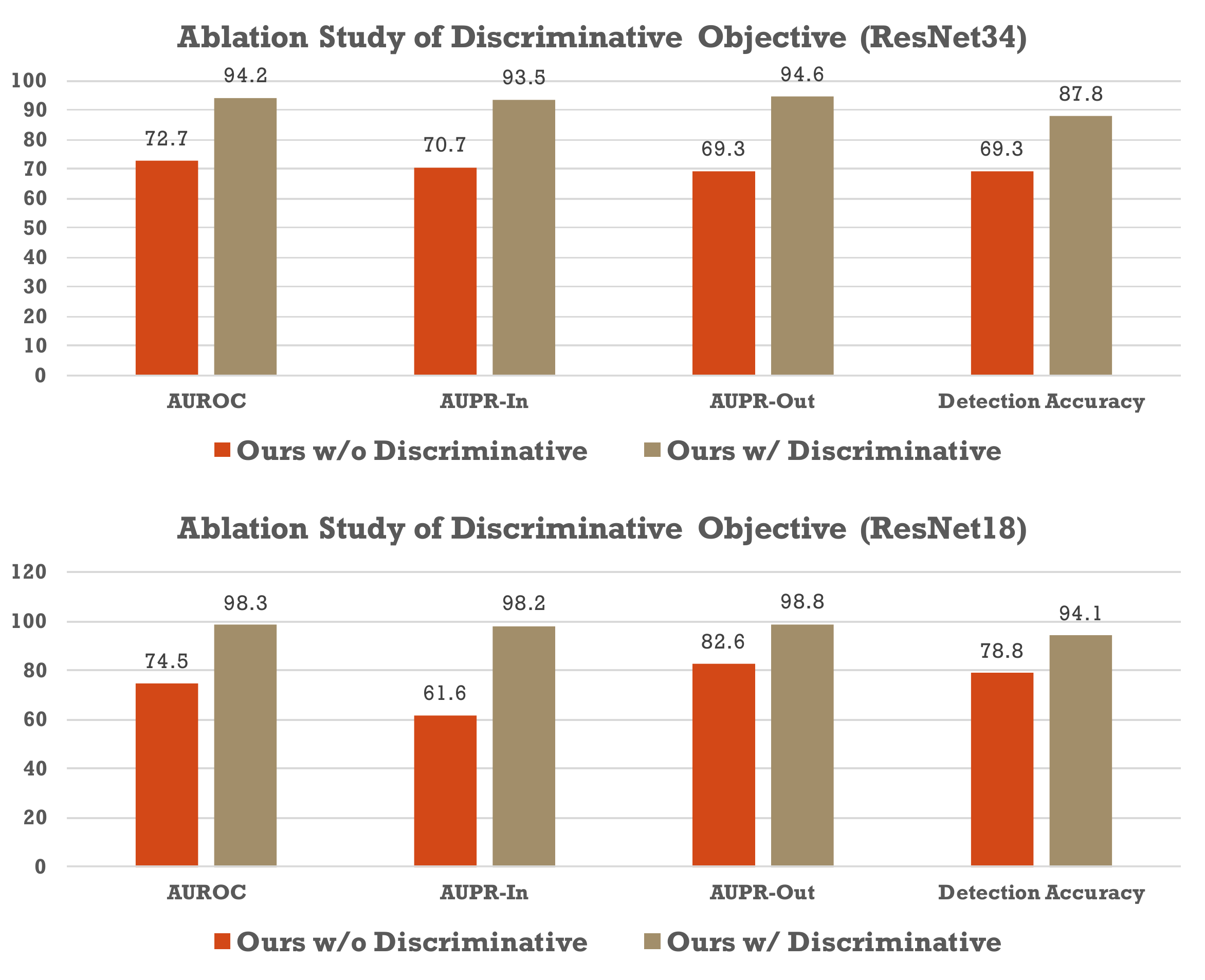}
    \caption{Impact of the discriminative objective on the final four detection metrics, the red part denotes without adopting the obejctive and khaki part denotes the increase adopted by adding the objective. }
    \label{fig:ablation_perturb}
\end{figure}
%\begin{figure}[thb]
%    \centering
%    \includegraphics[width=1.0\linewidth]{ablation_lambda.pdf}
%    \caption{Impact of the different balancing factor on the final AUROC detection metric under different architectures. }
%    \label{fig:ablation_lambda}
%\end{figure}
\paragraph{Analysis}
From the previous tables, we conclude that our method is able to generalize different architectures and datasets. The advantage of variational Dirichlet framework comes from its separation from uncertainty source and its efficiency to calculate the closed-form uncertainty measure. Combining with discriminative objective, the measure's robustness is further strengthened to provide better separation under more challenging tasks like CIFAR100.

\subsection{Ablation Study}
Here we perform ablation study to investigate the impacts of different settings (prior distribution, adversarial discriminative objective, balancing factor) with respect to the final evaluation metrics. 
\paragraph{Prior Distribution} Here we mainly experiment with four different prior distribution and depict our observations in~\autoref{fig:priors}. From which, we can observe that the non-informative uniform prior is a too strong assumption in terms of regularization, thus leads to inferior detection performances. In comparison, giving model one dimension of freedom can lead to generally better detection accuracy. Among these two priors, we found that preserving the groundtruth information can generally achieve slightly better performance, which is used through our experiments.    

\paragraph{Impact of Discriminative Objective}
In order to understand the effectiveness of our proposed discriminative function in out-of-distribution detection, we design parallel experiments to only train on the Variational Bayesian objective $\loss(\theta)$ and use the system for out-of-distribution detection. We plot the comparison results in~\autoref{fig:ablation_perturb}. From these two diagrams, we could observe a very significant increase across different metrics and network architectures. The other trend we observe is that our discriminative objective seems to yield lesser improvement on CIFAR10 than the CIFAR100 dataset.

\paragraph{Impact of Balancing Factor}
Here we design experiments to change the balancing factor from $0.001$ to $5$ to see its influence on the final detection accuracy metrics. We empirically observe that a small factor balancing factor only yields worse results, but a too large balancing factor will break down the algorithm. Fixing the factor at $0.1$ is able to achieve generally promising results on different datasets and architectures. 

\section{Related Work}
\subsection{Uncertainty Measure}
The novelty detection problem~\citep{pimentel2014review} has already a long-standing research topic in the traditional machine learning community, the previous works~\citep{vincent2003manifold,ghoting2008fast} have been mainly focused on low-dimensional and specific tasks. Their methods are known to be unreliable in high-dimensional space. Recently, more research works about detecting an anomaly in deep learning like \cite{akcay2018ganomaly} and \cite{lee2017training}, which propose to leverage adversarial training for detecting abnormal instances.  In order to make the deep model more robust to abnormal instances, different approaches like \cite{li2017learning} have been proposed to increase deep model's robustness against outlier during training. Another line of research is Bayesian Networks~\citep{gal2015bayesian,gal2016uncertainty}, which are powerful in providing stochasticity in deep neural networks by assuming the weights are stochastic. However, Bayesian Neural Networks' uncertainty measures like variational ratio and mutual information rely on Monte-Carlo estimation, where the networks have to perform forward passes many times, which is usually not practical in deep neural networks.
\subsection{Deep Latent Models}
In recent years, Bayesian inference has been widely used into deep learning to build deep latent models in many applications like computer vision and natural language processing like question answering~\citep{zhang2018variational,chen2018variational}, text generation~\citep{hu2017toward}, machine translation~\citep{zhang2016variational} and model compression~\cite{chirkova2018bayesian,chen2019large}. The use of deep latent model has greatly enhanced neural networks' capability to comprehend and model the uncertainty. In this paper, we are particularly interested in designing a new architecture based on Dirichlet distribution to better model the uncertainty in model's prediction.

\section{Conclusion}
In this paper, we aim at finding an effective way for deep neural networks to express their uncertainty over their output distribution. Our variational Dirichlet framework is empirically demonstrated to yield better results. The evaluation in terms of AUROC shows that the current method has already provided a very promising solution for small-scale out-of-distribution detection. In future work, the most interesting direction would be to apply such algorithm to even large-scale datasets like ImageNet or MSCOCO.

\bibliographystyle{named}
\bibliography{ijcai19}

\end{document}